# Thermal infrared image based vehicle detection in low-level illumination conditions using multi-level GANs

Shivom Bhargava[a], Sanjita Prajapati[a], Pranamesh Chakraborty[a,∗]

[a]*Department of Civil Engineering, Indian Institute of Technology Kanpur, Kanpur, U.P, India*

**Abstract**

Vehicle detection accuracy is fairly accurate in good-illumination conditions but susceptible to poor detection accuracy under low-light conditions. The combined effect of low-light and glare from vehicle headlight or tail-light results in misses in vehicle detection more likely by state-of-the-art object detection models. However, thermal infrared images are robust to illumination changes and are based on thermal radiation. Recently, Generative Adversarial Networks (GANs) have been extensively used in image domain transfer tasks. State-of-the-art GAN models have attempted to improve vehicle detection accuracy in night-time by converting infrared images to day-time RGB images. However, these models have been found to under-perform during night-time conditions compared to day-time conditions, as day-time infrared images looks different than night-time infrared images. Therefore, this study attempts to alleviate this shortcoming by proposing three different approaches based on combination of GAN models at two different levels that try to reduce the feature distribution gap between day-time and night-time infrared images. Quantitative analysis to compare the performance of the proposed models with the state-of-the-art models has been done by testing the models using state-of-the-art object detection models. Both the quantitative and qualitative analyses have shown that the proposed models outperform the state-of-the-art GAN models for vehicle detection in night-time conditions, showing the efficacy of the proposed models.

## 1. Introduction

Object detection algorithms have improved drastically over the past few years, both in terms accuracy of detection and frame rate (running time speed) (Zhao et al., 2019) and have been a popular area of research in the field of computer vision. Performance of the state-of-the-art object detection algorithms has been observed to improve significantly in daylight conditions using deep learning models. However, the scenario is drastically different under low illumination and poor weather conditions, where the performance of the state-of-the-art algorithms decreases significantly. The poor performance of object detection models in night-time conditions can be attributed to the glare from the headlights and tail-lights of vehicles, along with low illumination which lead to poor visibility of vehicle features. Consequently, object detection models frequently miss vehicles due to the intense glare or poor illumination. Therefore, there is a significant need for research and innovations in improving vehicle detection performance under low illumination and poor weather conditions.

Use of thermal infrared (TIR) cameras has been a popular research topic in Advanced Driving Assistance Systems (ADAS) to improve night-time perception of the driving environment. TIR cameras use thermal radiation and are therefore expected to be robust against illumination changes. Their robustness to illumination changes and shadow effects makes them useful in ADAS which is the key motivation of this research to use TIR cameras for improving vehicle detection performance under night-time conditions. One of the existing popular ways to use infrared images for vehicle detection purposes is to convert them into day-time RGB images, where vehicle detection already performs significantly well. This falls under image-translation task, which involves converting images from one domain (e.g., TIR image) to another (day-time RGB image). With the advancement in deep learning algorithms, a particular class of generative models, namely Generative Adversarial Networks (GANs), have become popular in generating images from random noises and also conditioned data, such as input images. The conditional GANs (CGANs) have been extensively used in image domain translation tasks (Isola et al., 2017). These models have been successful in synthesizing fake images which are realistic and like input images domain (Goodfellow et al., 2020). These conditional GANs are capable of image translation from one domain to the other given aligned data for supervised learning approaches (*e.g.*, Pix2pix) (Isola et al., 2017) and unaligned data for unsupervised learning based approaches (e.g., CycleGAN) (Zhu et al., 2017). Such CGANs have been used for variety of tasks such as semantic segmentation, grayscale visible image colorization, sketch to scene transformation, etc. Therefore, in this research, we have used conditional GANs to convert TIR images to day-time RGB to leverage robustness of infrared cameras to illumination conditions.

∗Corresponding author (pranames@iitk.ac.in)

Recent studies on converting TIR images to visual RGB conversion has been done using Convolution Neural Networks (CNN) by Berg et al. (2018) and improved further using GAN methodology by Kuang et al. (2020) in their model, named as Thermal Infrared Colorization using GANs or TIC-GAN. However, these existing GAN models such as CycleGAN, Pix2pix, and TIC-GAN have been trained and tested primarily on day-time images only. This is because the supervised models such as Pix2pix or TIC-GAN require an aligned dataset (day-time TIR and RGB images) to generate fake day-time RGB images. On the other hand, unsupervised models such as CycleGAN which don't require an aligned dataset cannot perform as good as the supervised models. Therefore, existing studies have used only day-time TIR images for image translation tasks. However, there is typically a distinct difference in the features in day-time and night-time TIR images, which can be also observed in Figure 1. Therefore, this leads to differences in training (day-time) and testing (night-time) data distribution, which can lead to poor performance of the GAN models in night-time conditions. This is because the supervised TIC-GAN model is trained using paired day-time RGB and infrared images. Therefore, in this study, we have proposed three models which uses the existing CycleGAN and TIC-GAN models as building blocks and attempts to reduce the training (day-time images) and test data (night-time images) feature distribution gap, which can help to improve the vehicle detection accuracies. We have tested our proposed approaches on the KAIST dataset (Hwang et al., 2015), both qualitatively and quantitatively using vehicle detection accuracies and compared their performance with the baseline TIC-GAN model.

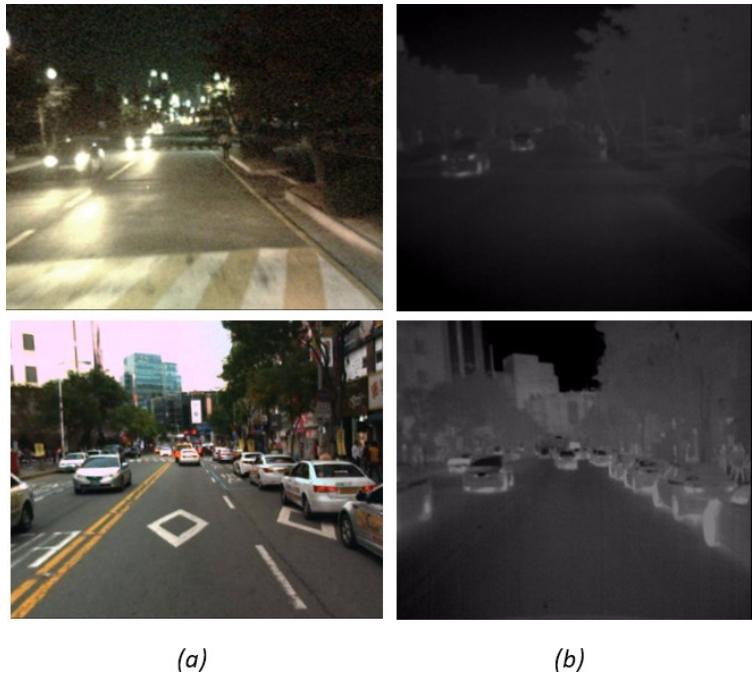

Figure 1: Difference in night-time and day-time images (a) RGB Ground truth (b) Infrared Image at night-time and day-time conditions

The following section presents the relevant literature and the motivation behind the study. In the third section implementation of models is done using existing studies and our proposed approaches are presented for the problem statement. The fourth section shows the results obtained using implementation of different models and proposed approaches along with discussion of the results. The fifth section concludes this study by featuring its contributions and identifying the future extensions.

## 2. Literature Review

In this research, the focus is to convert TIR images to day-time RGB using Generative Adversarial Networks (Goodfellow et al., 2020) for improving vehicle detection accuracy in scenario of low illumination conditions and glare coming from vehicles. Quite a few existing studies have been done already based on leveraging infrared images for vehicle detection.

Nam and Nam (2018) extracted vehicle features from infrared images in low illumination conditions using foreground extraction and classified vehicles based on vehicle-front features like headlights, grills regions, windshield, etc. using Bayesian classifier. However, a major drawback of the method in low-illumination condition is the low classification accuracy due to poor spatial resolution of thermal infrared-images to obtain fine texture details.

Ma et al. (2019) performed thermal infrared and visual grayscale image fusion, to differentiate targets and backgrounds based on difference in radiation. One of the major advantages of the method is that it can work competently in all-season and all-day and night conditions. Although the study was not focused on vehicle detection, it can be customized for vehicles as well. Wang et al. (2019) advanced the flexibility of image enhancement in poor-illumination images. The primary downside of the study is that it cannot be used to improve images based on video and real-time performance of the algorithm is poor.

Goodfellow *et al.* introduced Generative Adversarial Networks (GANs) (Goodfellow et al., 2020) in 2014 which was competent of producing new data instances that resembles the training data by varying input loss vector to generate changes in output features. GANs have been used extensively for image translation tasks, therefore researchers have also focused on using GANs for converting infrared images to RGB domain.

The GANs model optimizes itself such that the generated data distribution lies close to the input dataset distribution. This is done using two networks: the generator network ($G$) and the discriminator network ($D$). The generator network $G$ learns to upsample a random noise vector $z$ sampled from distribution $p_z$ to a realistic looking fake image $G(z)$. The model aims to bring $G(z)$ closer to the provided training dataset and this is achieved with the help of the discriminator network $D$ which is a classifier network that tries to distinguish between real sample $x$ and fake sample $G(z)$ and scores the output based on realness of the output $G(z)$ it produces. This score is called the generative adversarial loss. The generator network tries to maximize this loss whereas the discriminator tries to minimize it. This *minimax* game ultimately drives the generator network to produce high fidelity fake images $G(z)$ and the generated data distribution $p_g$ eventually comes close to training data distribution $p_{data}$.

The adversarial loss function is expressed as shown below:

$$\min_G \max_D V(D, G) = \mathbb{E}_{x \sim p_{\text{data}}(x)}[\log D(x)] + \mathbb{E}_{z \sim p_z(z)}[\log(1 - D(G(z)))] \tag{1}$$

However, the output of basic GANs network has a major drawback that it cannot generate desired class of output as the noise vector is random. This problem can be addressed using conditional GANs (Mirza and Osindero, 2014). In conditional GANs, the generator and discriminator network are trained on a supervised learning approach. Pix2pix GANs (Isola et al., 2017), is one of the earliest conditional GANs model for image translation tasks. The supervised data contains images from two domains having perfect pixel wise correspondence. However, this model also requires pixel wise correspondence, which is difficult to obtain between RGB and infrared images. Also feature-wise comparison of generated result and target ground truth is missing in Pix2pix which can lead to lack of fidelity in generated result.

CycleGAN (Mirza and Osindero, 2014) is one of the conditional GAN models based on unsupervised learning. It was introduced by Zhu *et al* in 2017. It is useful for the image translation task where supervised dataset is not available, and hence it does not require a target output and learns about the style and content from the target domain to generate fake images. To achieve unsupervised image translation, CycleGAN uses image reconstruction by including cycle consistency loss. It also includes an adversarial loss for generating realness in the generated outputs. CycleGAN consists of two generator discrimination network pairs in series to put an additional constraint of mapping in both forward and backward directions. Both the generator networks and both the discriminator networks are exactly similar in architecture. The discriminator network is the PatchGAN discriminator network. PatchGAN, focus on penalizing image patches separately rather than penalizing complete image at once.

Recently, Thermal Infrared colorization GAN or TIC-GAN was proposed by Kuang et al. (2020). TIC-GAN leverages supervised data for the training process. It produces better results by including feature-based comparison rather than only focusing on pixel-to-pixel comparison between target and generated images. To address the infrared colorization problem, TIC-GAN used a feature-based loss function called perceptual loss. This helps model in developing feature-based understanding of the domains and the generated images can be expected be more robust to small pixel wise changes. Also, it includes total variational loss which results in significant noise reduction in generated output. TIC-GAN has a generator network to perform infrared to visual RGB domain translation and a discriminator network to optimize the generator network's performance. The TIC-GAN model uses an additional VGG-16 network for feature-based comparison of generated and target images *i.e.* called perceptual loss. The model uses the Pix2pix HD (Wang et al., 2018) generator network to obtain high resolution generated images of superior quality. The discriminator network in TIC-GAN is like the CycleGAN model *i.e.* the PatchGAN discriminator.

ToDayGAN (Anoosheh et al., 2019) was introduced by Asha Anoosheh in 2019. It is used for day-night localization tasks with the help of daytime reference set and nighttime query set. In ToDayGAN, an image translation model is trained to translate between day and night image domain. Then, night-to-day direction is used to transform night-time images into day-time images and both translated images and reference images are then fed to a featurization process to get a feature vector. Nearest neighbor search is then used to get the closest matching reference image. The image translation model is built using ComboGAN (Anoosheh et al., 2018), while the generator network is identical to the CycleGAN.

## 3. Methodology

Vehicle detection accuracy on RGB images often drops substantially in combined case of low illumination conditions and glare coming from vehicles during night-time conditions. In such situations, thermal infrared images which only sense thermal radiation can be thought to be used as a feasible solution. One approach can be to convert infrared image to day-time RGB image to overcome the problem of low-illumination and glare from vehicles simultaneously. To implement the conversion of infrared to RGB, GANs can be a suitable choice as it has been extensively used in image domain translation tasks.

Converting infrared image to its corresponding day-time RGB is an interesting task, because unlike grayscale image to RGB conversion which requires only estimation of chrominance *i.e.* color, infrared to RGB conversion requires estimation of both chrominance and luminance. Popular conditional GAN models like Pix2pix (see Figure 20 of (Isola et al., 2017)) and CycleGAN perform poorly in the above-mentioned task since getting both unknown entities *i.e.* chrominance and luminance may also require feature-based understanding. The TIC-GAN model, on the other hand, holds the capability of converting day-time infrared input to day-time RGB images. However, the model was not found to perform well in converting night-time infrared images to their corresponding day-time RGB images. The primary reason behind this being the gap in training input (day-time infrared) and testing input (night-time infrared) image feature distribution.

A major difference in feature distributions of infrared images at night-time and day-time is the contrast conditions. The contrast conditions in the infrared images can be found to be better in the day-time than the night-time infrared images (Berg et al., 2018). In simple terms, the surroundings during night-time adopt a more homogeneous temperature, thus making contrast in the thermal infrared images in the night to be lower than the day-time infrared images. Therefore, reducing the feature distribution gaps between training and testing data is required, to keep the test performance as per the training performance.

To achieve this, we have proposed three different methodologies. Our proposed approaches use the advantages of existing GAN models, such as CycleGAN, ToDayGAN, and TIC-GAN. In all three approaches, the test input is night-time infrared images, and we are using TIC-GAN for producing the final output, that is day-time RGB images. The type of data used for training TIC-GAN is different in all three approaches and attempts to reduce the gap in training and testing data distribution to improve the test results qualitatively and quantitatively, presented in the subsequent subsections. Training details of proposed approaches are explained in the "Dataset and Training Details" subsection at the "Results and Discussion" section.

### 3.1. Proposed approach-1

In this approach, we are generating day-time RGB using day-time TIR images. Efforts have been made to reduce the training and testing input distribution gap by making test data input like the training data input. The test input which is night-time TIR Image can be converted into day-time infrared domain, because TIC-GAN model performs well on day-time infrared input.

The training of TIC-GAN model has been done on day-time TIR with corresponding day-time RGB paired dataset in supervised training. The test input has also been converted to fake day-time TIR using CycleGAN for this unsupervised domain translation task and used as test input for TIC-GAN.

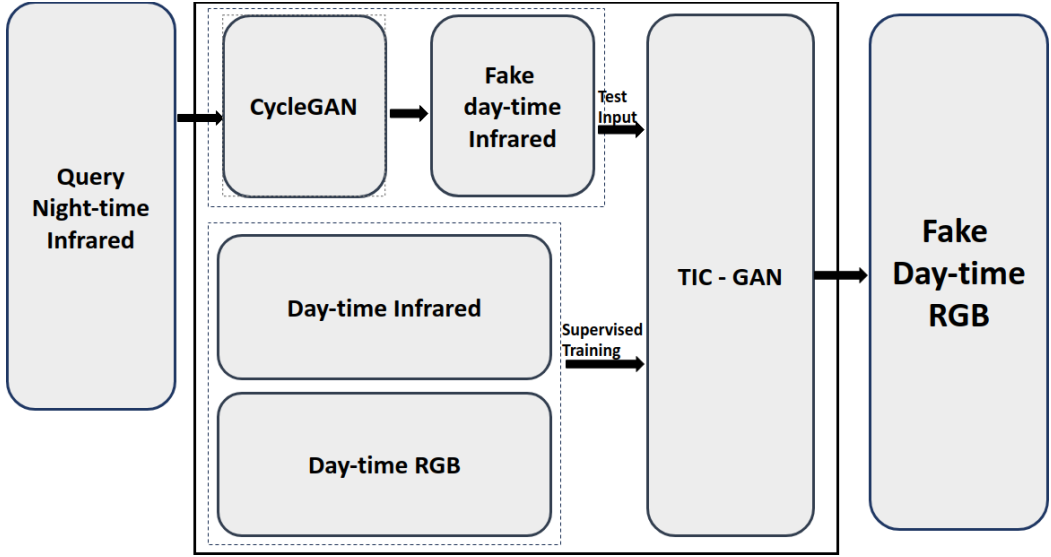

Figure 2: Pipeline for approach 1

The pipeline for the network is shown in the Figure 2 which shows that query night-time TIR image is first converted to day-time TIR domain image using CycleGAN. This fake day-time TIR image is served as test input to the TIC-GAN model which is trained on day-time TIR with corresponding day-time RGB aligned dataset. The TIC-GAN model when given the fake day-time TIR input finally generates the fake day-time RGB image.

*3.2. Proposed approach-2*

In this approach, we are generating day-time RGB from night-time TIR. To keep training and testing input from same domain, I.e., night-time TIR, we can convert training input from night-time RBG to day-time RGB.

This approach keeps training and testing input to be from the same domain *i.e.* night-time infrared unlike proposed approach-1 in which test input was modified to be consistent with training input. For converting night-time infrared to day-time RGB this approach uses ToDayGAN as one of the components along with TIC-GAN.

ToDayGAN (Anoosheh et al., 2019) is based on unsupervised domain translation for night-time RGB images to day-time RGB. To reduce the feature gap between training and testing input, it can be a better idea to train and test on the same kind of dataset *i.e.* the training and testing should both be done on night-time infrared images. Since we want to convert night-time infrared input to day-time RGB and we want to use night-time infrared images in the training dataset, we do not have a corresponding day-time RGB ground truth. To alleviate that, we have leveraged ToDayGAN to convert the night-time RGB to fake day-time RGB ground truth. Ultimately, this can provide night-time infrared images for both training and testing, and the TIC-GAN model can be trained with night-time infrared and corresponding day-time RGB obtained using ToDayGAN.

Using this, we can synthesize supervised dataset consisting of night-time infrared and corresponding synthesized fake day-time RGB. Therefore, the TIC-GAN model can be trained directly on night-time infrared and corresponding day-time RGB, testing can be done on night-time infrared. The pipeline of the proposed approach is shown by Figure 3.

*3.3. Proposed approach-3*

In this approach, we can modify training data input (day-time infrared image) to look similar to testing data input domain (night-time infrared image). To convert day-time infrared to fake night-time infrared, CycleGAN (unsupervised domain translation) has been used. The resulting fake night-time infrared is paired with the corresponding ground truth day RGB. Training of TIC-GAN model is then done on synthesized night-time infrared (synthesized from day-time infrared input using CycleGAN) and corresponding ground truth day-time RGB, and finally we test the TIC-GAN model on the night-time infrared query image. This can help to bring the feature distributions of infrared images closer to each other during training and testing.

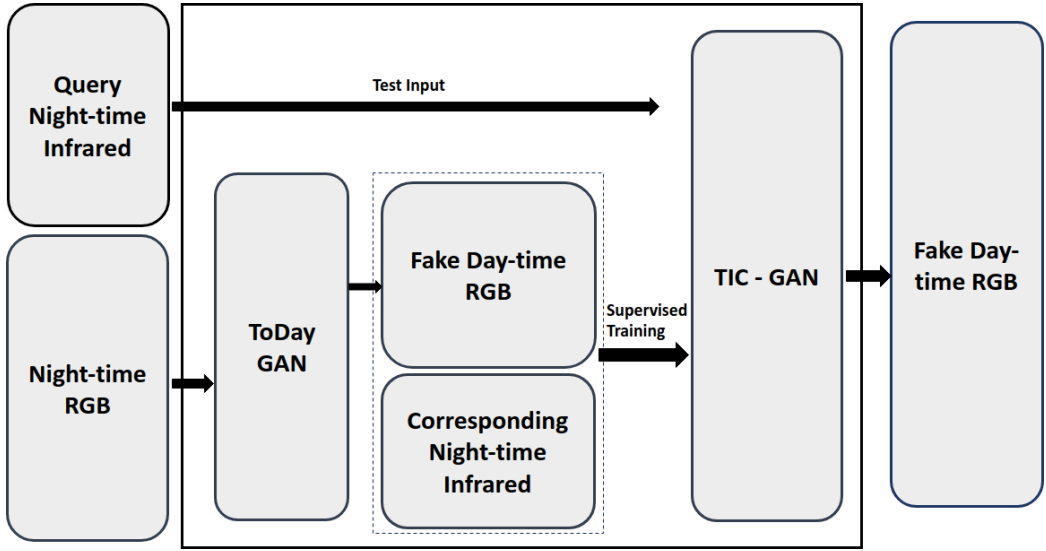

Figure 3: Pipeline for approach 2

The idea is to bring query day-time infrared images as close as possible to night-time infrared. The better this conversion would be, the more realistic results we will be getting. CycleGAN model is used for the conversion of day-time infrared to night-time infrared conversion. The higher the quality of conversion of day-time infrared to night-time infrared, the better will be the quality of results that will be obtained at the test time from input night-time infrared using the TIC-GAN model.

Figure 4 shows the pipeline for the proposed approach-3 for better visualization of the approach. In the next section, we discuss in detail the results obtained from various proposed approaches and the baseline TIC-GAN model, their shortcomings, qualitative and quantitative evaluation of the proposed approaches and TIC-GAN.

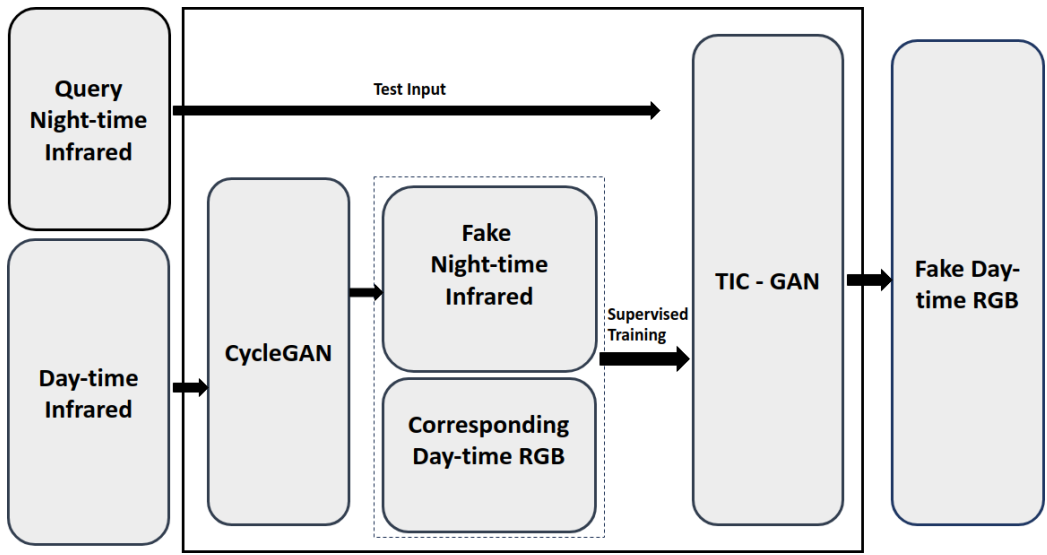

Figure 4: Pipeline for approach 3

## 4. Results and discussions

In this section, we provide the detailed results obtained from the three proposed approaches in this study along with the comparison with the TIC-GAN model as the baseline algorithm. We first discuss the details of the dataset on which the algorithms have been trained and tested, along with the training details of each model. Then we show the qualitative comparison of the models, followed by the quantitative comparison using object detection performance metrics.

*4.1. Dataset and Training Details*

All experiments in this study have been performed using the KAIST thermal dataset (Hwang et al., 2015). This dataset consists of 95k thermal-RGB pairs (640x480, 20Hz) taken from a vehicle. Out of this, about 66% of the image pairs are of day-time and rest are night-time images.

In proposed approach-1, CycleGAN model is trained to transform query night-time infrared image to day-time infrared. And for doing so, it is trained on the night-time infrared and day-time infrared domains. We used 5000 images each from both day-time and night-time infrared domains. The model is trained for 60 epochs, with a learning rate of 0.0002 for the first 40 epochs which is further decayed to 0 over the next 20 epochs. Rest all the hyperparameters are set to their default values as mentioned in (Zhu et al., 2017). Also, TIC-GAN model is used in proposed approach-1 for converting infrared to RGB, and it is trained on KAIST paired thermal-RGB dataset with 5000 image pairs. The model is trained for 100 epochs, 80 epochs with a learning rate $\alpha=0.0002$ which is gradually reduced to 0 in the next 20 epochs. Other aspects and hyperparameters while training is exactly similar to the values used in the original TIC-GAN paper (Kuang et al., 2020).

In proposed approach-2, for training ToDayGAN model KAIST dataset is used in unaligned manner with good illumination night-time images in one domain and day-time images in the other domain. Training was done for 80 epochs. Learning rates are started at 2e-4 for generator networks and 1e-4 for discriminator networks and are kept constant for the first training half and linearly decreased to zero in the second half of training. $\lambda$ is the hyperparameter for cycle consistency loss is set to 10.0. Also, TIC-GAN model is used in proposed approach-2 for converting night-time infrared to day-time RGB. For doing so, corresponding day-time RGB for night-time infrared is obtained by using ToDayGAN by converting night-time RGB to day-time RGB. This synthesized dataset having night-time infrared and synthesized day-time RGB consists of 600 image pairs. The model is trained for 100 epochs, 80 epochs with a learning rate $\alpha=0.0002$ which is gradually reduced to 0 in the next 20 epochs. Other aspects and hyperparameters while training were exactly similar to the values used in the original TIC-GAN paper.

In proposed approach-3, CycleGAN is trained to transform day-time infrared to night-time infrared. For which it is trained on the day-time infrared and night-time infrared domains. We used 5000 images each from both day-time and night-time infrared domains. The rest of the training settings for CycleGAN are exactly similar to the CycleGAN model used in proposed approach-1. TIC-GAN model is trained to transform night-time infrared to day-time RGB. For training TIC-GAN, we use fake night-time infrared obtained from CycleGAN model which is paired with day-time RGB. The Synthesized dataset contains 5000 image pairs. Other training settings for TIC-GAN are exactly like the proposed approach-1 TIC-GAN model.

*4.2. Qualitative Analysis*

In this study, three different approaches have been proposed to reduce the feature distribution gap between the night-time TIR and day-time TIR while training and testing to improve night-time TIR to day-time RGB conversion. Figure 5 shows the sample results of the images obtained from the baseline TIC-GAN, proposed approach-1, proposed approach-2, and proposed approach-3, and the ground truth RGB images. It can be seen from the figure that the proposed approach-3 performs better than the TIC-GAN and the other proposed approaches. Significant blurring is present in the outputs from TIC-GAN and proposed approaches 1 and 2, which can make vehicle detection difficult. Further, some artifacts such as pink patches were also observed in outputs from the proposed approach 2, as shown in Figure 4d, top row. Please note that the baseline TIC-GAN in this study has been tested on night-time images, unlike the study by Kuang et al. (2020), where TIC-GAN was proposed and tested on day-time images. Since the focus of this study is to improve vehicle detection in night-time conditions, testing on the night-time infrared images shows that the shortcomings of TIC-GAN under such low illumination conditions, where proposed approaches, in particular proposed approach 3 perform significantly better.

In addition to overall qualitative evaluation of the quality of the generated images, we also need to evaluate the models' performance with respect to vehicle detection. Vehicle detection using object detection algorithms *e.g.* YOLOv5

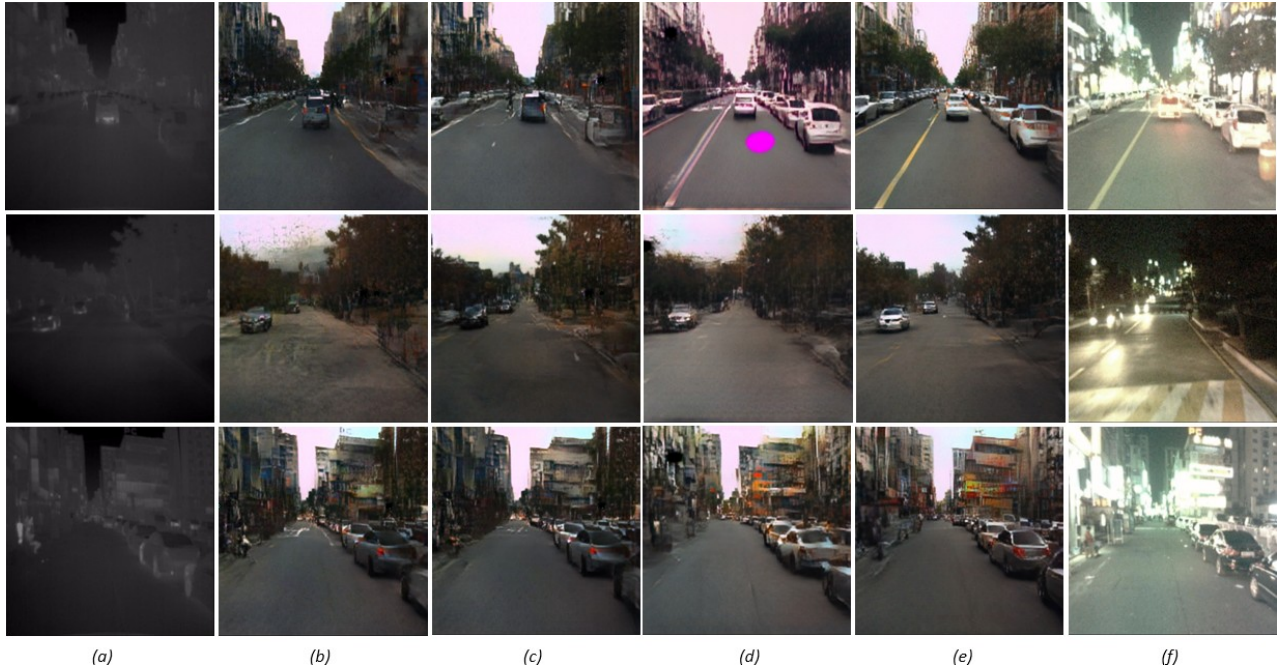

Figure 5: Qualitative comparison of the TIC-GAN and proposed approaches: (a) Night-time Infrared (b) TIC-GAN (c) Proposed approach-1 (d) Proposed approach-2 (e) Proposed approach-3 (f) Ground Truth RGB

(You Only Look Once) (Redmon et al., 2016) is reliably accurate in day-time conditions. However it is vulnerable to low accuracy in low-light conditions due to noisy images, insufficient illumination to reveal vehicle features and also due to glare from vehicle headlights and tail-lights towards the camera.

Figure 6 shows the comparison of object detection results from the groundtruth night-time RGB, TIC-GAN, and the proposed approaches. The top two rows show images from good illumination conditions due to street lights, while the bottom two rows are from images under poor illumination conditions. It is evident from the figure that the low-light environment is challenging for vehicle detection and vehicles are highly vulnerable to be omitted for detection (false negatives), due to combined effect of glare from vehicles and low illumination conditions which make vehicle detection omission more likely. In contrast, the proposed approach-3 shows considerable reduction in such false negatives *i.e.* lesser omission of vehicles due to low-illumination and glare. Further, the artifacts such as pink patches observed in outputs from proposed approach 2 also found to not impact the vehicle detection performance.

Under good illumination conditions (Figure 6a top two rows), we can observe that object detection in case of night-time RGB have fewer false negatives (vehicle omission in detection). This can be attributed to the fact that even in case of glares from vehicle headlights, the vehicle features are still sufficiently visible due to good illumination conditions. Also, conversion from night-time infrared to day-time RGB often produces noisy patches in outputs which can in turn increase the vulnerability for higher false positive rate (false alarm for object detection) than that of object detection on good illumination night-time images. Nonetheless, it can be observed from Figure 6 that proposed approach-3 has lower false positives (false vehicle detection) and lower false negatives (vehicle detection omission) as compared to TIC-GAN, proposed approach-1, proposed approach-2 under poor illumination conditions. Next, we discuss the quantitative comparison of the proposed approaches using vehicle detection accuracies.

*4.3. Quantitative Evaluation*

From the results described above, it can be observed that the proposed approach-3 works best in case of night-time TIR to day-time RGB translation task qualitatively, particularly under low illumination night-time conditions. Quantitatively, this is not an easy task for comparison since groundtruth day-time RGB images are not available for comparison with the generated day-time images, which can be used for pixelwise comparison. Hence, in this study, we have used object detection metrics as a quantitative comparison criteria instead. We have used the state-of-the-art You Only Look Once (YOLOv5) model (Redmon et al., 2016), trained on COCO (Common Objects in Context) dataset for object detection on the generated images. These test metrics can serve as a quantitative comparison criteria of results generated using different approaches.

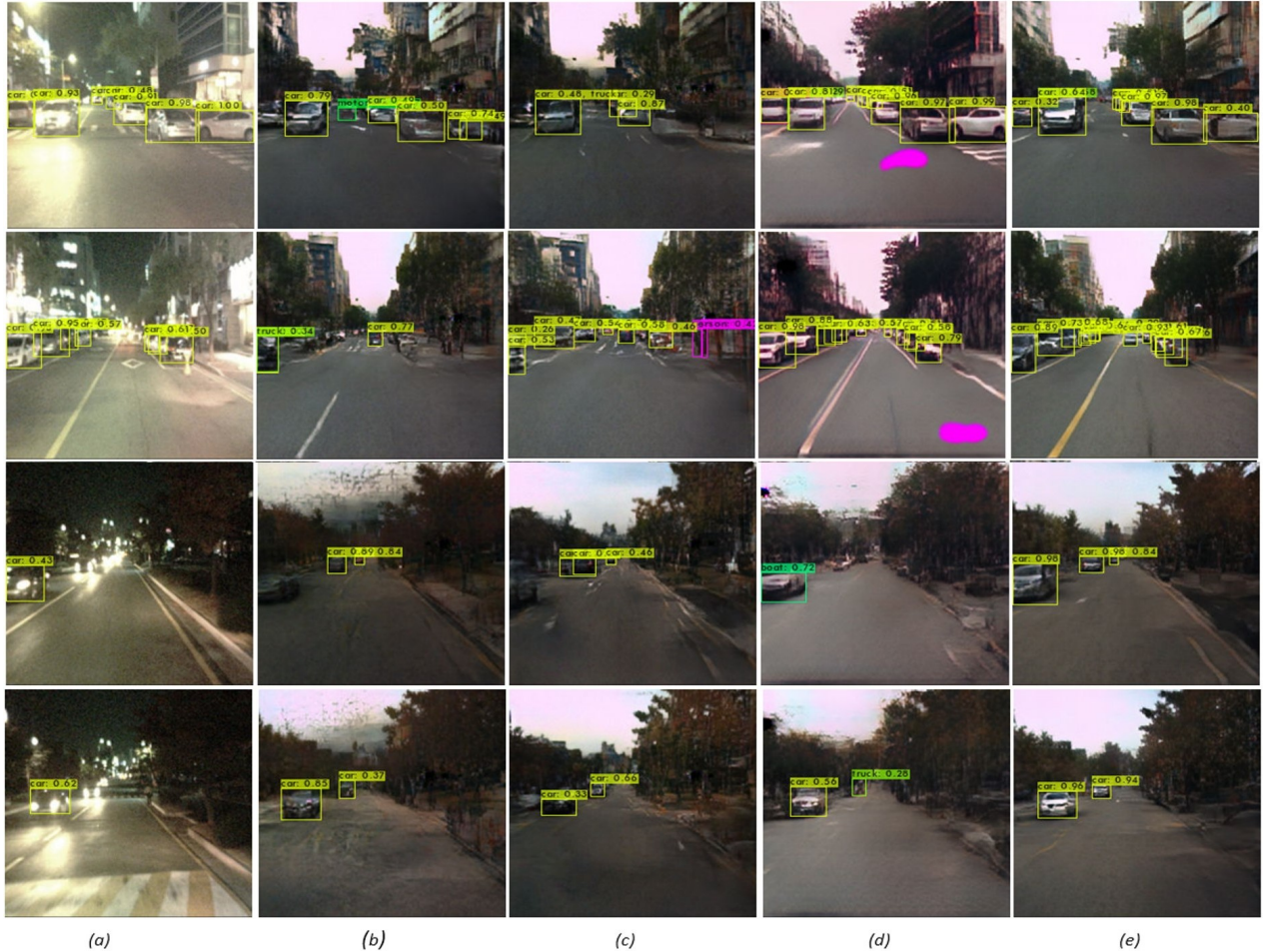

Figure 6: Object detection results comparison: (a) Groundtruth Night-time RGB (b) TIC-GAN (c) Proposed approach-1 (d) Proposed approach-2 (e) Proposed approach-3

All vehicles (cars, buses, trucks, motorcycles, and bicycles) were manually annotated in 200 ground truth night-time RGB images to create the test dataset for vehicle detection comparison. These annotated vehicles were then compared with the same vehicle classes detected using YOLOv5 on the ground-truth RGB images and the images generated using TIC-GAN and the proposed three approaches. Further, a separate test dataset containing 200 day-time images has been annotated and tested directly using YOLOv5, which can be used as the performance criteria, that the proposed approaches should ideally achieve. The groundtruth test RGB images are sampled randomly from the KAIST dataset, used in this study.

Table 1 shows the precision, recall, and mean Average Precision@0.5 on the generated results obtained using the three different proposed approaches, TIC-GAN, corresponding ground-truth night-time RGB, and the day-time RGB for the test dataset. The number of images on which each model is tested along with the number of vehicles that were present in the images and annotated are also mentioned in the Table 1. Further, the precision-recall (PR) curve, obtained from this test dataset, is shown in Figure 7. Here, *AP* or the average precision is the area under PR-curve and mean of *AP*'s for all categories of vehicles on which the model is trained to detect gives the mAP. The mean average precision values are obtained at Intersection over Union threshold of 0.5 also called as "mAP@0.5".

From the Figure 7, it can be seen that the PR-Curve of the three proposed approaches outperforms the state- of-the-art TIC-GAN model. This is also observed from the precision, recall, and *mAP@*0.5 values shown in Table 1. In particular, proposed approach 3 performs better compared to the TIC-GAN and the remaining two proposed

Table 1: Object detection evaluation results

| Model | Images | Annotated Vehicles | Precision | Recall | mAP@0.5 |
|---|---|---|---|---|---|
| **TIC-GAN** | 200 | 987 | 0.563 | 0.342 | 0.347 |
| **Proposed approach-1** | 200 | 987 | 0.658 | 0.348 | 0.441 |
| **Proposed approach-2** | 200 | 987 | 0.734 | 0.480 | 0.571 |
| **Proposed approach-3** | 200 | 987 | 0.681 | 0.614 | 0.667 |
| **Night-time RGB** | 200 | 987 | 0.861 | 0.653 | 0.797 |
| **Day-time RGB** | 200 | 1247 | 0.877 | 0.881 | 0.907 |

approaches, which was also observed based on the qualitative evaluation results.

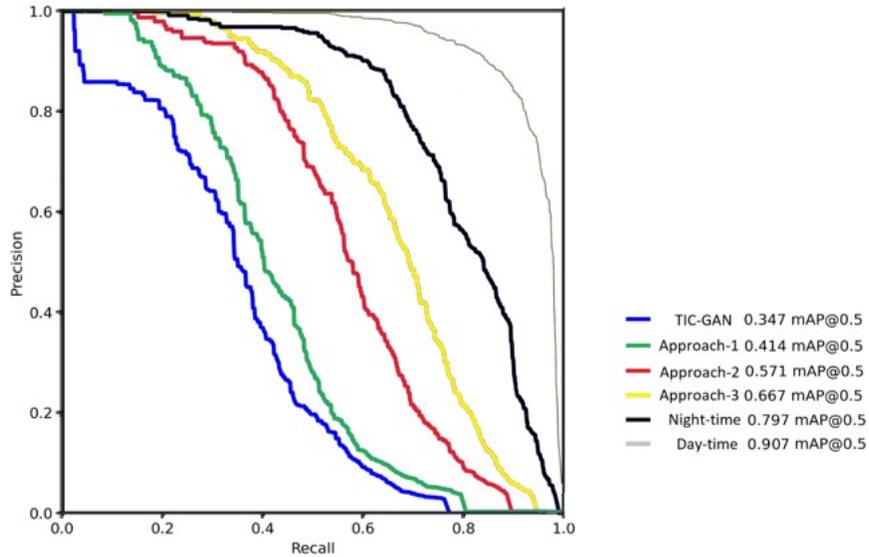

Figure 7: Precision-Recall curve on object detection results

However, *mAP@*0.5 and the Precision-Recall curve obtained from the groundtruth night-time RGB images are still found to be better than the baseline and all the proposed approaches. One of the possible reason behind this observation can be that the majority of night-time images in KAIST dataset (Kuang et al., 2020) contains good illumination night-time images. On the other hand, most of the low illumination images in KAIST dataset had empty streets and very few vehicles were present in such images which made the test dataset and consequently the analysis more biased towards the good illumination night-time images. Figure 8 shows some sample low illumination images with empty streets, along with the output of fake day-time image obtained using proposed approach 3. Further, the bottom two rows of images shown in Figure 6 shows the performance of different models under low illumination conditions, where the proposed approach-3 can be seen to perform better than the baseline TIC-GAN model.

Therefore, it can stated that the proposed approach-3 has shown good results in night-time infrared to day-time RGB conversion task and its use is specifically important for perceiving night-time environment in case of low illumination conditions. It can be seen that the model is robust towards poor illumination conditions and we can leverage this property for better perception of dark environment in advanced driving assistance systems and surveillance applications.

## 5. Conclusion

In this research, we have proposed three different approaches to handle low accuracy of vehicle detection in dark environments or poor-illumination conditions using thermal infrared images. Glare from vehicles and low-illumination conditions combined results in poor vehicle detection performance. However, infrared images are robust to illumination changes and dependent on thermal conditions of the environment. Keeping this in mind, to address this problem of improving vehicle detection accuracy in low-light conditions, we have converted night-time infrared images to day-time RGB images using GAN networks. GAN models such as TIC-GAN, which has been used in literature for conversion of infrared to RGB images, have been found to perform well in day-time images, but their performance drops significantly in night-time conditions. This is because the supervised TIC-GAN model is trained using paired day-time RGB and infrared images. Therefore, in this study, we have proposed models which uses the existing CycleGAN and TIC-GAN models as building blocks, and attempts to reduce the training (day-time images) and test data (night-time images) feature distribution gap, which can help to improve the vehicle detection accuracies.

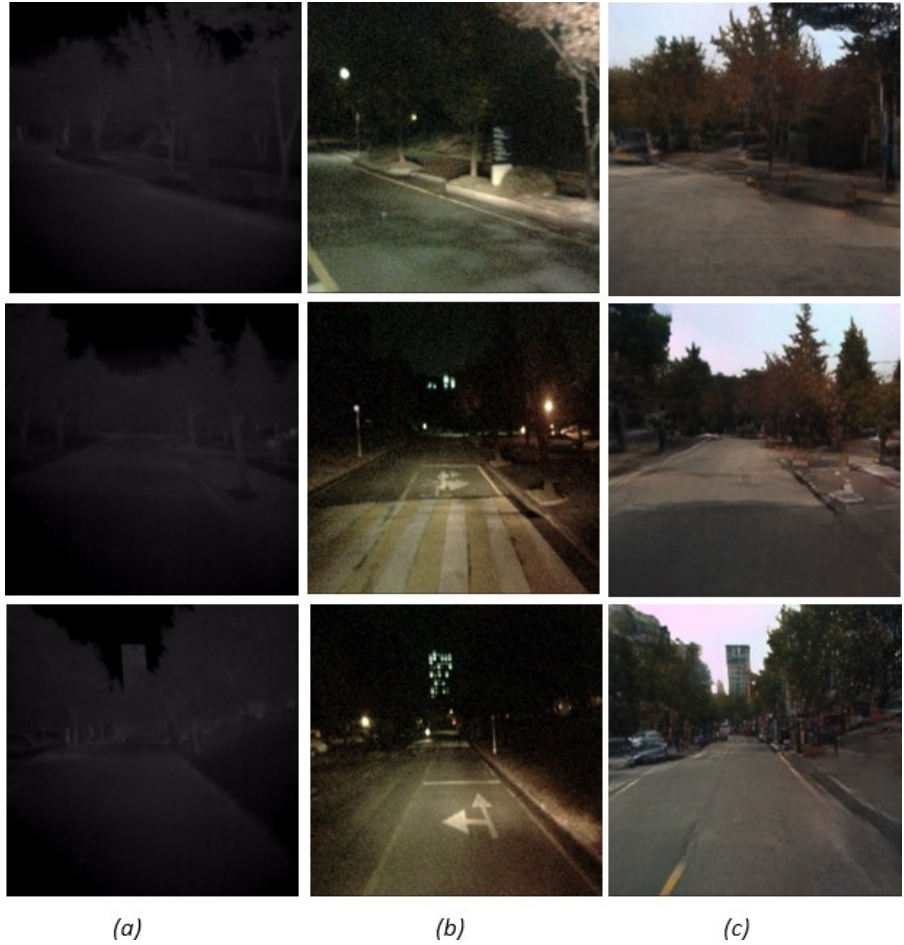

Figure 8: Dark environment perception in night-time : (a) Night-time Infrared (b) Ground truth RGB (b) Output from Proposed approach-3

The proposed approach-1 first converts the query night-time infrared image to day-time infrared image using CycleGAN which forms the input to the TIC-GAN model to output fake day-time RGB image. On the other hand, proposed approach-2 converts night-time RGB to fake day-time RGB using ToDayGAN, which is used to train the TIC-GAN model using fake daytime RGB and corresponding night-time infrared image. The training data of TIC-GAN model in the proposed approach-3, however consist of fake night-time infrared (obtained using cycleGAN) and corresponding day-time RGB image. These attempts help to reduce the training and test data distribution gap. We test our proposed approaches on the KAIST dataset, both qualitatively and quantitatively using vehicle detection accuracies and compare their performance with the baseline TIC-GAN model. Our analyses show that the proposed approach-3 outperforms the baseline and the other proposed approaches. Significantly less artifacts and blurring is observed in the outputs obtained using the proposed approach-3 and the model performs well even in the challenging low-illumination night-time conditions.

However, the object detection evaluation metrics of night-time RGB has been found to be higher than the proposed approach-3 based detection. This can be due to the sparse representation of poor illumination based night-time traffic images in the KAIST dataset, which also has mostly empty streets with very few vehicles. Due to this fact, majority of analysis is more biased towards good illumination condition based night-time traffic scenes, where existing object detection models already performs fairly well. Thus, it can be concluded that while object detection performs fairly well in good illumination night-time conditions, however their performance drops significantly under low-time illumination conditions where the proposed approach-3 can be used based on thermal infrared images to improve vehicle detection accuracies. In future, the proposed approaches can be trained and tested on large-scale self-curated challenging low illumination images to further determine the efficacy of the models. Further, models based on fusion of RGB and thermal infrared images, both in training and inferencing time, can also be looked upon to improve the vehicle detection accuracies under all illumination conditions.

## 6. Acknowledgements

Our research results are based upon work supported by the Initiation Grant scheme of Indian Institute of Technology Kanpur (IITK/CE/2019378). Any opinions, findings, and conclusions or recommendations expressed in this material are those of the author(s) and do not necessarily reflect the views of the IITK.